\documentclass[10pt,twocolumn]{IEEEtran}
\usepackage{amsfonts,subfigure,multicol,multirow,color,verbatim,graphicx,cite,epsfig,amssymb,amsmath}
\usepackage{tabularx}
\usepackage{booktabs}
\usepackage{subfigure}
\usepackage{tabularx}
\usepackage{booktabs}
\usepackage{subfigure}
\usepackage{graphicx}
\usepackage{textcomp}
\usepackage{multirow}
\begin{document}
\title{ICDAR2019 Competition on Scanned Receipt OCR and Information Extraction}

\author{\normalsize Zheng~Huang, Kai~Chen, Jianhua~He, Xiang~Bai, Dimosthenis~Karatzas, Shjian~Lu, and C.V.~Jawahar
\thanks{Z. Huang is with Shanghai Jiaotong University, China, huang-zheng@sjtu.edu.cn. 
Jianhua~He is with Aston University, UK,  j.he7@aston.ac.uk.
Kai Chen (kaichen@onlyou.com) is with Onlyou, China.
Xiang Bai (xbai@hust.edu.cn) is with Huazhong University of Science and Technology, China.
Dimosthenis Karatzas (dimos@cvc.uab.es), Universitat Autónoma de Barcelona, Spain.
Shijian Lu (Shijian.Lu@ntu.edu.sg) is with Nanyang Technological University, Singapore. 
C. V. Jawahar (jawahar@iiit.ac.in) is with IIIT Hyderabad, India.
}}

\maketitle

\begin{abstract}
Scanned receipts OCR and key information extraction (SROIE) represent the processeses of recognizing text from scanned receipts
and extracting key texts from them and save the extracted tests to structured documents.
SROIE plays critical roles for many document analysis applications and holds great commercial potentials,
but very little research works and advances have been published in this area.
In recognition of the technical challenges, importance and huge commercial potentials of SROIE,
we organized the ICDAR 2019 competition on SROIE.
In this competition, we set up three tasks, namely,
Scanned Receipt Text Localisation (Task 1), Scanned Receipt OCR (Task 2) and Key Information Extraction from Scanned Receipts (Task 3).
A new dataset with 1000 whole scanned receipt images and annotations is created for the competition.
The competition opened on 10th February, 2019 and closed on 5th May, 2019.
There are 29, 24 and 18 valide submissions received for the three competition tasks, respectively. 
In this report we will presents the motivation, competition datasets, task definition, evaluation protocol, submission statistics,
performance of submitted methods and results analysis. 
According to the wide interests gained through SROIE and the healthy number of submissions from academic, research institutes and industry over different countries,
we believe the competition SROIE is successful. 
And it is interesting to observe many new ideas and approaches are proposed for the new competition task set on key information extraction. 
According to the performance of the submissions, we believe there is still a large gap on the expected information extraction performance. 
Task of key information extraction is still very challenging and can be set for many other important document analysis applications.
It is hoped that this competition will help draw more attention from the community and promote research and development efforts on SROIE. 

\end{abstract}

\section{INTRODUCTION}\label{sec:introduction}

Scanned receipts OCR is a process of recognizing text from scanned structured and semi-structured receipts and invoices. On the other hand, extracting key texts from receipts and invoices and save the texts to structured documents can serve many applications and services, such as efficient archiving, fast indexing and document analytics. Scanned receipts OCR and key information extraction (SROIE) play critical roles in streamlining document-intensive processes and office automation in many financial, accounting and taxation areas. However, SROIE also faces big challenges. With performance greatly boosted by recent breakthroughs in deep learning technologies in terms of accuracy and processing speed, OCR is becoming mature for many practical tasks (such as name card recognition, license plate recognition and hand-written text recognition). However, receipts OCR has much higher accuracy requirements than the general OCR tasks for many commercial applications. And SROIE becomes more challenging when the scanned receipts have low quality. Therefore, in the existing SROIE systems, human resources are still heavily used in SROIE. There is an urgent need to research and develop fast, efficient and robust SROIE systems to reduce and even eliminate manual work.    

With the trends of OCR systems going to be more intelligent and document analytics, SROIE holds unprecedented potentials and opportunities, which attracted huge interests from big companies, such as Google, Baidu and Alibaba. Surprisingly, there are little research works published in the topic of SROIE. While robust reading, document layout analysis and named entity recognition are relevant to the SROIE, none of the existing research and past ICDAR competitions fully address the problems faced by SROIE [1,2,3]. 

In recognition of the above challenges, importance and huge commercial potentials of SROIE, we organized the ICDAR 2019 competition on SROIE, aiming to draw attention from the community and promote research and development efforts on SROIE. We believe the this competition could be of interests to the ICDAR community from several aspects.

First, to support the competition, a large-scale and well-annotated invoice datasets are provided, which is crucial to the success of deep learning based OCR systems. While many datasets have been collected for OCR research and competitions, to the best of our knowledge, there is no publicly available receipt dataset. Compared to the existing ICDAR and other OCR datasets, the new dataset has some special features and challenges, e.g., some receipts having poor paper quality, poor ink and printing quality; low resolution scanner and scanning distortion; folded invoices; too many unneeded interfering texts in complex layouts; long texts and small font sizes. To address the potential privacy issue, some sensitive fields (such as name, address and contact number etc) of the receipts are blurred. The datasets can be an excellent complement to the existing ICDAR and other OCR datasets.

Second, three specific tasks are proposed: receipt OCR and key information extraction. Compared to the other widely studied OCR tasks for ICDAR, receipt OCR (including text detection and recognition) is a much less studied problem and has some unique challenges. On the other hand, research works on extraction of key information from receipts have been rarely published. The traditional approaches used in the named entity recognition research are not directly applicable to the second task of SROIE. 

Third, comprehensive evaluation method is developed for the two competition tasks. In combination with the new receipt datasets, it enables wide development, evaluation and enhancement of OCR and information extraction technologies for SROIE. It will help attract interests on SROIE, inspire new insights, ideas and approaches.

The competition opened on 10th February, 2019 and closed on 5th May, 2019.
There are 29, 24 and 18 valide submissions received for the three competition tasks, respectively. 
In this report we will presents the motivation, competition datasets, task definition, evaluation protocol, submission statistics,
performance of submitted methods and results analysis. 
According to the wide interests gained through SROIE and the healthy number of submissions from academic, research institutes and industry over different countries,
we believe the competition SROIE is successful. 
It is hoped that this competition will help draw more attention from the community and promote research and development efforts on SROIE. 

\section{Dataset and Annotations}
The dataset has 1000 whole scanned receipt images, which are used for all three competition tasks but with different annotations. 
Each receipt image contains around about four key text fields, such as goods name, unit price and total cost, etc. The text annotated in the dataset mainly consists of digits and English characters. Figure 1 shows some example scanned receipt images.

The dataset is split into a training/validation set (“trainval”) and a test set (“test”). The “trainval” set consists of 600 receipt images are made available to the participants along with their annotations. The “test” set consists of 400 images, which will be made available a few weeks before the submission deadline.

\begin{figure}[!htb]
\centering		
	\subfigure[Task 1 and Task 2.]{
		\centering
		\includegraphics[width=80mm]{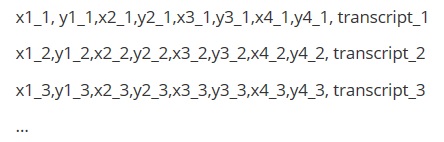}
		\label{fig:annotation-ocr}
	} 	
	\subfigure[Task 3.]{
		\centering
		\includegraphics[width=90mm]{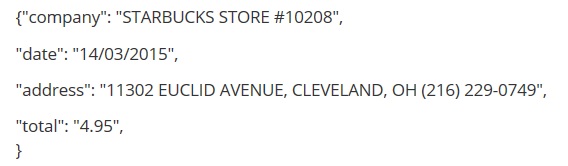}
		\label{fig:annotation-ie}
	} 			
\caption{Annotation format for the tasks. }
\label{fig:ann-format}	 		
\end{figure}

For receipt detection and OCR tasks, each image in the dataset is annotated with text bounding boxes (bbox) and the transcript of each text bbox. Locations are annotated as rectangles with four vertices, which are in clockwise order starting from the top. Annotations for an image are stored in a text file with the same file name. The annotation format is similar to that of ICDAR2015 dataset, which is illustrated in Fig.\ref{fig:annotation-ocr}. 
For the information extraction task, each image in the dataset is also annotated and stored with a text file with format, which is illustrated in Fig.\ref{fig:annotation-ie}.

\begin{figure*}[!htb]
\centering		
	\subfigure[]{
		\centering
		\includegraphics[width=30mm]{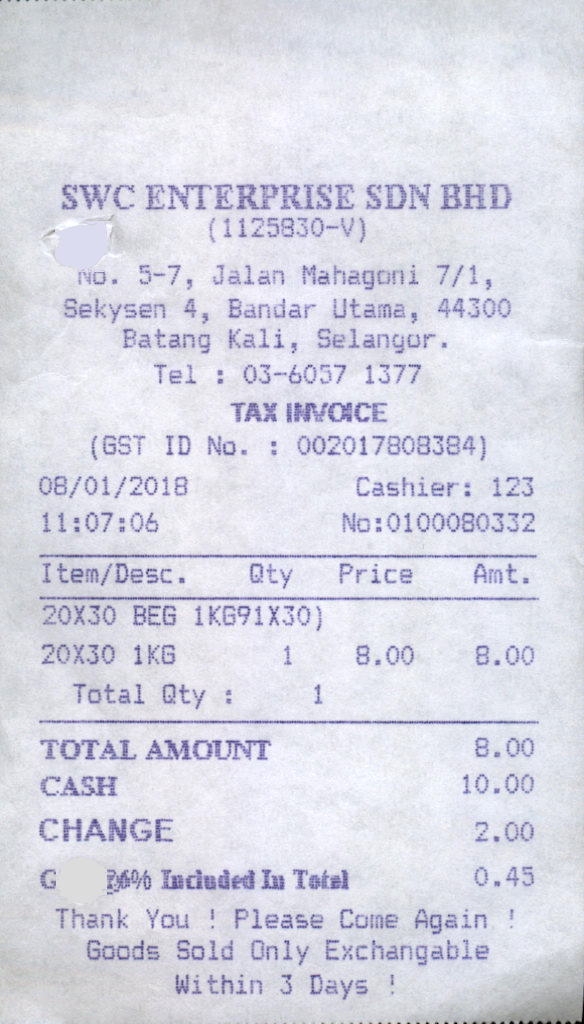}
		\label{fig:net-example1}
	} 	
	\subfigure[]{
		\centering
		\includegraphics[width=30mm]{X51005268408.jpg}
		\label{fig:net-example2}
	} 			
	\subfigure[]{
		\centering
		\includegraphics[width=30mm]{X51006387931.jpg}
		\label{fig:net-example3}
	} 	
	\subfigure[]{
		\centering
		\includegraphics[width=30mm]{X51006388068.jpg}
		\label{fig:net-example4}
	} 	
	\subfigure[]{
		\centering
		\includegraphics[width=30mm]{X51006619564.jpg}
		\label{fig:net-example5}
	} 		
\caption{Examples of scanned receipts for the competition tasks. }
\label{fig:net-capacity}	 		
\end{figure*}

\section{Competition Tasks}

\subsection{Task 1 - Scanned Receipt Text Localisation}
\subsubsection{Task Description}

Localizing and recognizing text are conventional tasks that has appeared in many previous competitions, such as the ICDAR Robust Reading Competition (RRC) 2013, ICDAR RRC 2015 and ICDAR RRC 2017 [1][2]. The aim of this task is to accurately localize texts with 4 vertices. The text localization ground truth will be at least at the level of words. Participants will be asked to submit a zip file containing results for all test images.

\subsubsection{Evaluation Protocol}

As participating teams may apply localization algorithms to locate text at different levels (e.g. text lines), for the evaluation of text localizazation in this task, the methodology based on DetVal will be implemented. The methodology address partly the problem of one-to-many and many to one correspondences of detected texts. In our evaluation protocol mean average precision (mAP) and average recall will be calculated, based on which F1 score will be computed and used for ranking [3]. 

\subsection{Task 2 - Scanned Receipt OCR}

\subsubsection{Task Description}

The aim of this task is to accurately recognize the text in a receipt image. No localisation information is provided, or is required. Instead the participants are required to offer a list of words recognised in the image. The task will be restricted to words comprising Latin characters and numbers only.

The ground truth needed to train for this task is the list of words that appear in the transcriptions. In order to obtain the ground truth for this task, you should tokenise all strings splitting on space. For example the string “Date: 12/3/56” should be tokenised “Date:”, “12/3/56”. While the string “Date: 12 / 3 / 56” should be tokenised “Date:” “12”, “/”, “3”, “/”, “56”.

\subsubsection{Evaluation Protocol}
For the Recognition ranking we will match all words provided by the participants to the words in the ground truth. If certain words are repeated in an image, they are expected to also be repeated in the submission results. We will calculate Precision (\# or correct matches over the number of detected words) and Recall (\# of correct matches over the number of ground truth words) metrics, and will use the F1 score as the final ranking metric.

\subsection{Task 3- Key Information Extraction from Scanned Receipts}
\subsubsection{Task Description}
The aim of this task is to extract texts of a number of key fields from given receipts, and save the texts for each receipt image in a json file. Participants will be asked to submit a zip file containing results for all test invoice images. 

\subsubsection{Evaluation Protocol}
For each test receipt image, the extracted text is compared to the ground truth. An extract text is marked as correct if both submitted content and category of the extracted text matches the groundtruth; Otherwise, marked as incorrect. The mAP is computed over all the extracted texts of all the test receipt images. F1 scored is computed based on mAP and recall. F1 score is used for ranking.

\section{Organization}

The competition SROIE made use of the Robust Reading Competition (RRC) web portal to maintain information of the competition, download links for the datasets, and user interfaces for participants to register and submit their results [4]. 
Great support has been received from the RRC web team. 
The timeine for the SROIE competition is shown below.
\begin{itemize}
\item Website online: February 15, 2019
\item Registration open: February 20 – March 31, 2019
\item Training and validation datasets available: March 1, 2019
\item Test dataset available: March 31, 2019
\item Tasks 1 and 2 submission deadline: April 22, 2019
\item Task 3 submission open: April 23, 2019
\item Task 3 submission Deadline: May 5, 2019
\end{itemize}

By the submission deadline, we received  29 submissions  for Task 1, 24 for Task 2 and 18 for Task 3.
Some teams submitted multiple results for the same task. 
We took measures to identify multiple submissions and only took the last submission from mulitple ones in ranking.

\section{Submission and Results}

After the submission deadlines, we collected all submissions and evaluate their performance through automated process
with scripts developed by the RRC web team. 
The winners are determined for each task based on the score achieved by the corresponding primary metric.
A workshop is planned to present the summary of the competition and give awards to the winners.

\subsection{Task 1 Performance and Ranking}
\begin{figure*}[!htb]
\centering		
	\includegraphics[width=150mm]{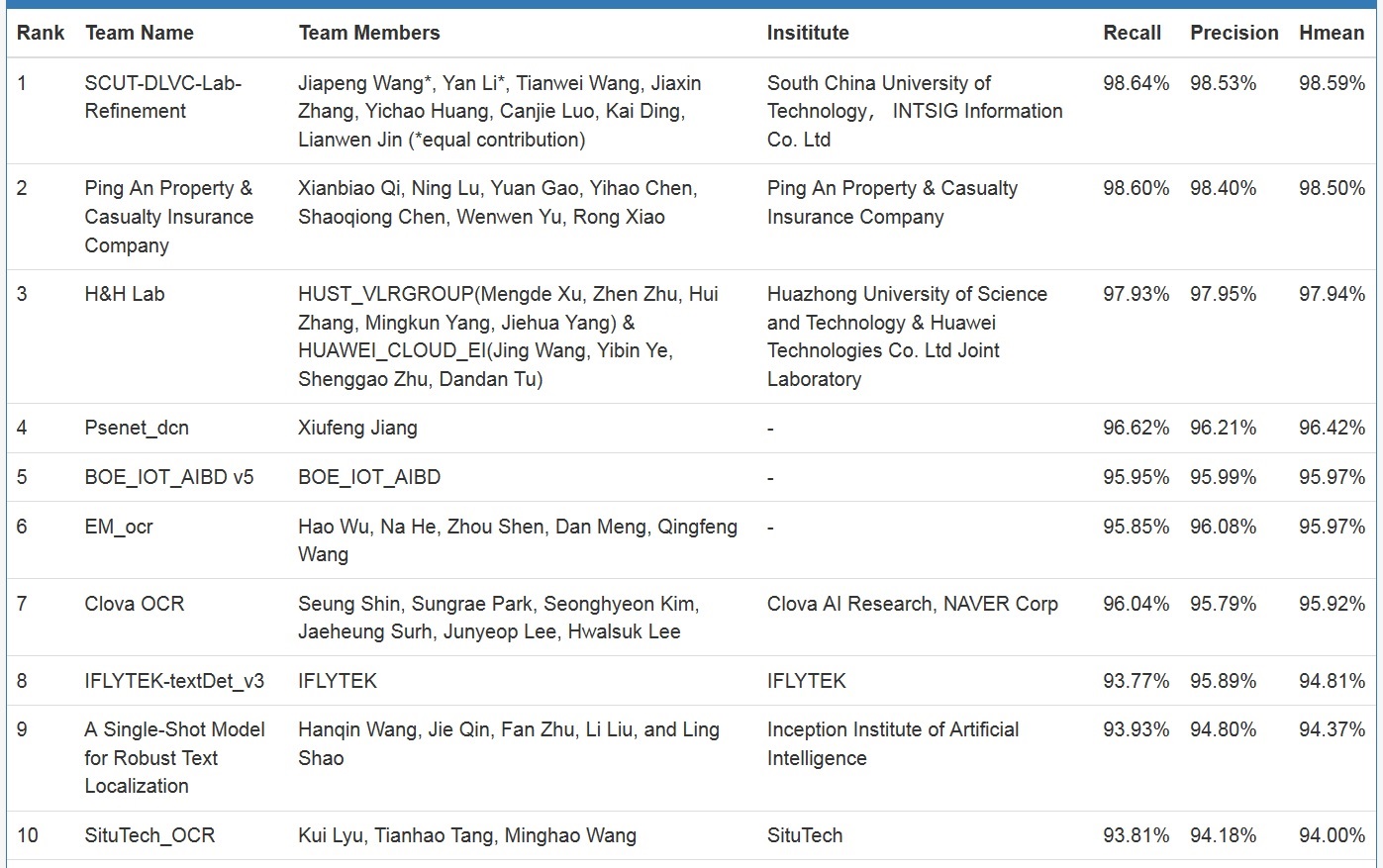}
\caption{Top 10 methods for Task 1 - Scanned Receipt Text Localisation. }
\label{fig:task1-ranking} 		
\end{figure*}

Fig.~\ref{fig:task1-ranking} lists the top-10 submissions of Task 1. 

The methods used by the top 3 submissions for Task 1 are presented below.
\begin{enumerate}
\item 1st ranking method ``SCUT-DLVC-Lab-Refinement'': This method utilizes a refinement-based Mask-RCNN. In the method redundant information and refine detected bounding-box results are iteratively removed. Finally, the submitted result is ensembled from several models with different backbones.
\item 2nd ranking method: ``Ping An Property \& Casualty Insurance Company'': This method uses an anchor-free detection framework with FishNet as the backbone. 
The training and testing images are pre-processed to roughly align their scales with OpenCV adaptive threshold. 
The detection results are post-processed according to the classification score map. It can well solve the problem of short detection of long text line.
\item 3rd ranking method `` H\&H Lab'':  In this method EAST and multi oriented corner are ensembled to create a robust scene text detector. 
To make network learning easier, the mutli-oriented corner network is modified with a new branch borrowed from east added.
\end{enumerate}

\subsection{Task 2 Performance and Ranking}
\begin{figure*}[!htb]
\centering		
	\includegraphics[width=150mm]{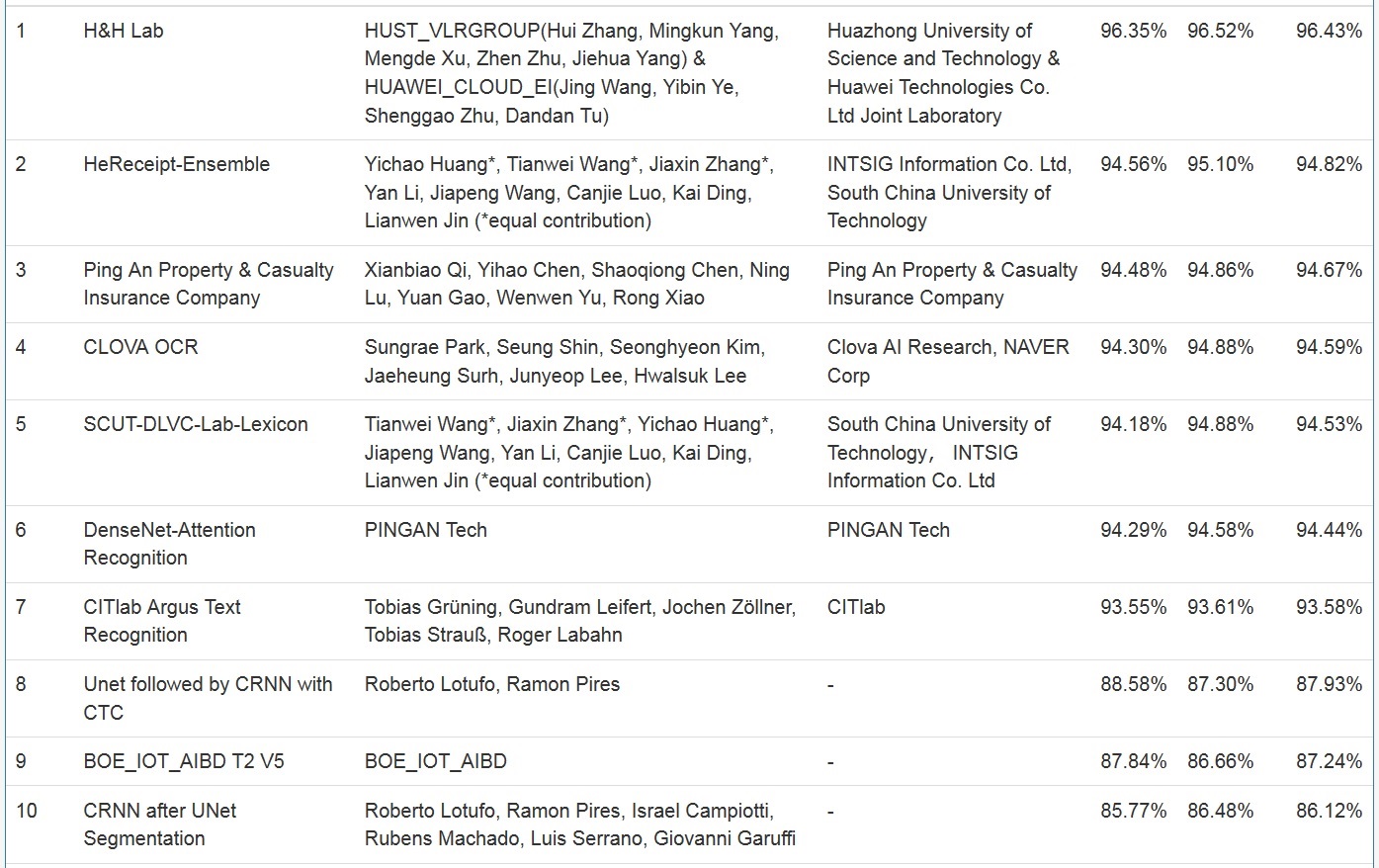}
\caption{Top 10 methods for Task 2 - Scanned Receipt OCR. }
\label{fig:task2-ranking}	 		
\end{figure*}

Fig.~\ref{fig:task2-ranking} lists the top-10 submissions of Task 2. 

The methods used by the top 3 submissions for Task 2 are presented below.
\begin{enumerate}
\item 1st ranking method `` H\&H Lab'': This method mainly uses CRNN for Task 2. Different from the content in the paper, the structure of CNN is modifified to be PVANet-like and uses multiple GRU layers. The training strategy is adjusted to further improve the recognition result.
\item 2nd ranking method ''INTSIG-HeReceipt-Ensemble'':  This recognition method is based on CNN and RNN. With different backbones and recurrent structure settings, the team trained several models seperately and then implement model ensemble. Finally, according to the official requirement, they split line output into word level.
\item 3rd ranking method ``Ping An Property \& Casualty Insurance Company'': The team employed an encoder-decoder sequence method with attention mechanism. First, they created 2 millions of systhesis text lines, where the receipt background is used. Each line consists of one to five words. Then, they fine-tuned the network with real-world receipt data.
\end{enumerate}

\subsection{Task 3 Performance and Ranking}
\begin{figure*}[!htb]
\centering		
	\includegraphics[width=150mm]{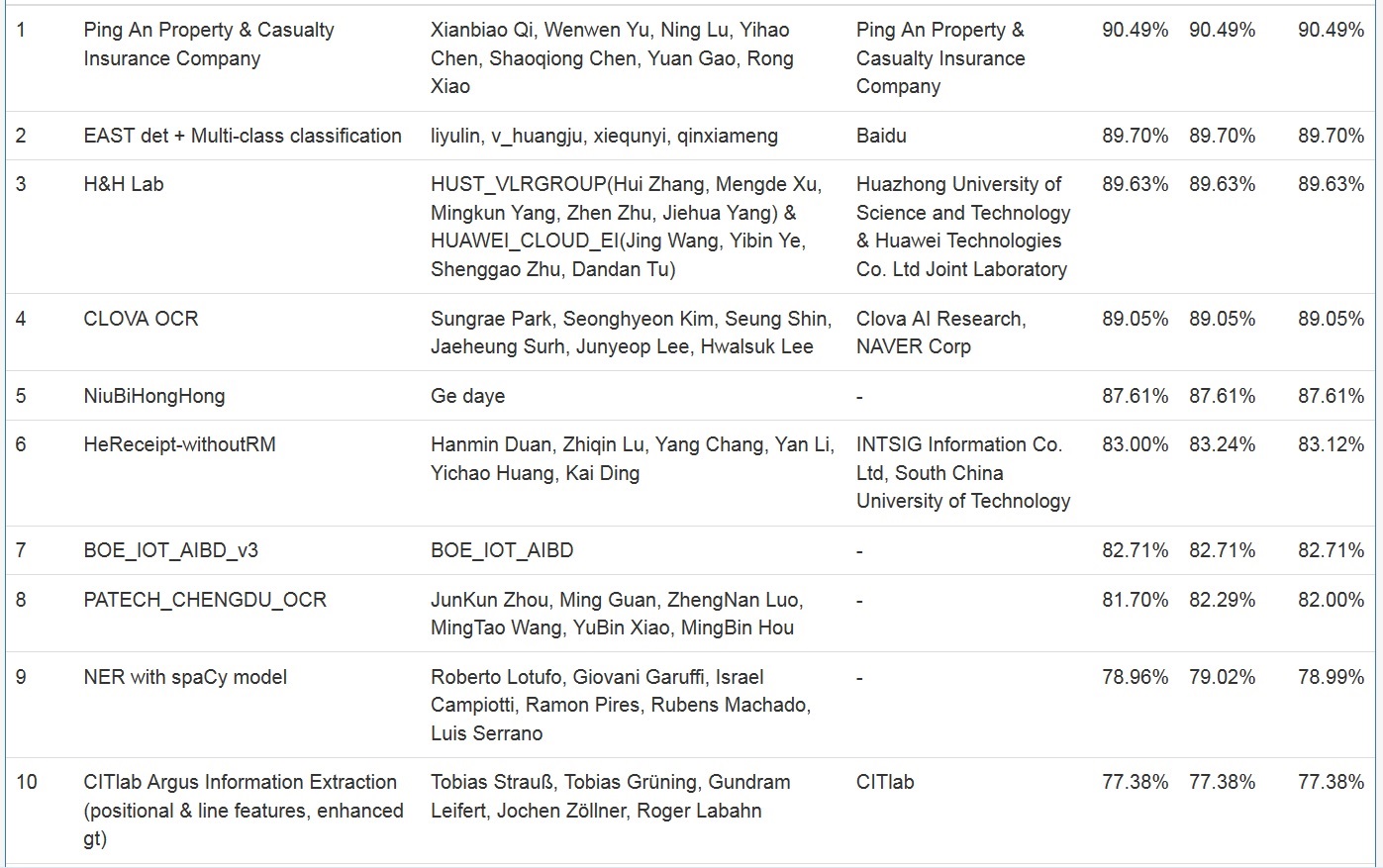}
\caption{Top 10 methods for Task 3 - Key Information Extraction from Scanned Receipts. }
\label{fig:task3-ranking} 		
\end{figure*}

Fig.~\ref{fig:task3-ranking} lists the top-10 submissions of Task 3. 

The methods used by the top 3 submissions for Task 3 are presented below.
\begin{enumerate}
\item 1st ranking method ``Ping An Property \& Casualty Insurance Company'': Based on our detection and recognition results on Task1 and 2, this team used a lexicon (which is built from the train data set ) to auto-correct results and use RegEx to extract key information. For every keyword (company, cash, date, and address), they used different patterns of regular expression.
\item 2nd ranking method: ``Enetity detection'': This method combines the strategy of content parsing and entity awared detector based on `EAST` to obtain selected candidates. A text classifier with RNN embedding was applied for final results.
\item 3rd ranking method `` H\&H Lab'':  The team completed Task 3 using a structure similar to BiGRU-CNNs-CRF, and designed a number of constraint rules to correct the results.
\end{enumerate}

\section{Discussions}

As both OCR and information extraction of scanned receipts are new tasks for ICDAR competitions,
the organizers proposed to set the tasks of detection and OCR for the SROIE competition along side the task of information extraction.
There are two main considerations: firstly, we hope to attract sufficient interest for this new competition with some established taskes; secondly,
text detection and OCR are key processes for the task of information extraction.
According to the submitted methods and results, we believe such setting on the comptetion tasks is successful.
The number of submissions for all the three tasks are encouragingly high. And the top ranking teams for Task 3 are mostly doing well with the first two tasks.

From the aspects of task performance, unsurprisingly, the detection and OCR results for Task 1 and Task 2 are very good, with 16 teams obtaining Hmean of more than 90\% for Taks 1 and 7 teams obtained Hmean of more than 90\% for Task 2. The results for Taks 1 and Task 2 shows that the existing approaches for general text detection and recognition are performing well for scanned receipts. However, if we take the strict requirement of receipt applications into account, say 99\% accuracy, it is noted that, even the best OCR method in Task 2 can't deliver the required performance. Therefore, we believe that the challenge for scanned receipt OCR is still there, which demands further research and development efforts.
For the Task 3 of key information extraction, we can see only one method achieves Hmean of more than 90\% (which is 90.49\%), and more than half of the submitted methods achieve Hmean of less than 80\%. Therefore, there are still large space for improvement for the new task of key information extraction.

From the aspects of technical approaches used in the subitted methods, it is observed that for the first two tasks, while the traditional network models are applied, many top ranking methods use ensamble of multiple architectures or models to improve performance. And some methods use large datasets of synthetic texts. 
For the Task 3 of key information extraction, it is interesting to observe many new ideas and approaches are proposed. Most of the submitted methods use different ideas and approaches. As this task is new and presents an open research issues, we expect more innovative approaches will be proposed after the competion.

\section{Conclusion}

We organized the one of the first competitions on the OCR and information extraction for scanned receipts.
For the competition SROIE we prepared new datasets and evaluation protocols for three competition tasks. 
A good number of submissions were received for all three tasks,
which showed a broad interests on the topic from the academic and industry.
And it is interesting to observe many new ideas and approaches are proposed for the new competition task of key information extraction. 
According to the performance of the submissions, we believe
there is still a large gap on the expected information extraction performance. 
The task of key information extraction is still very challenging and can be set for many other important document analysis applications.
It will be interesting to extend on this competition with more challenging and larger datasets and applications in the future.  
The new datasets used in this competition will be made available after the event.

\section*{Acknowledgement}
The ICDAR19 competition SROIE is supported in part by the European Union’s Horizon 2020 research and innovation programme under the Marie Skłodowska-Curie grant agreement No 824019. 
The organizers thank Cheng Li of Onlyou for his great support on the creation of the competition datasets and overall managment,
Sergi Robles and the RRC web team for their tremendous support on the registration, submission and evaluation jobs.

\vfill


\begin{thebibliography}{10}

\bibitem{He10}
He J., Chen H., \textit{et al}, `Adaptive congestion control for DSRC vehicle networks',
\textit{IEEE Comm. Lett.}, Feb. 2010, \textbf{14}, (2), p.127-129.

\bibitem{Kar13}
D. Karatzas, F. Shafait, S. Uchida, M. Iwamura, L. Gomez, S. Robles, J. Mas, D. Fernandez, J. Almazan, L.P. de las Heras: ICDAR 2013 Robust Reading Competition. ICDAR, 2013.
\bibitem{Kar15}
D. Karatzas, L. Gomez-Bigorda, A. Nicolaou, D. Ghosh , A. Bagdanov, M. Iwamura, J. Matas, L. Neumann, VR. Chandrasekhar, S. Lu, F. Shafait, S. Uchida, E. Valveny: ICDAR 2015 robust reading competition. ICDAR, 2015.
\bibitem{Ev15}
Everingham, M. and Eslami, S. M. A. and Van Gool, L. and Williams, C. K. I. and Winn, J. and Zisserman, A.: The Pascal Visual Object Classes Challenge: A Retrospective. IJCV, 2015
\bibitem{Kar19}
D. Karatzas, L. Rushinol, The Robust Reading Competition Annotation and Evaluation Platform.



\end{thebibliography}
\end{document}